\documentclass[sigconf,anonymous]{acmart}

\AtBeginDocument{%
  }

\setcopyright{none}
\renewcommand\footnotetextcopyrightpermission[1]{}

\usepackage{amsmath}
\usepackage{booktabs}
\usepackage{graphicx}
\usepackage[table]{xcolor}
\usepackage{tabularx}
\usepackage{placeins}
\usepackage{cuted}
\usepackage{subcaption}

\makeatletter
\def\@subsubsecfont{\bfseries\section@raggedright}
\makeatother
\graphicspath{{fig/}}

\def\sysname{EventRadar}
\newcommand{\sage}{SAGE}
\newcommand{\chg}{CHG-LISTA}
\newcommand{\R}{\mathbb{R}}
\DeclareMathOperator*{\argmax}{arg\,max}

\makeatletter
\def\@mkauthors{%
  \global\setbox\mktitle@bx=\vbox{%
    \noindent\unvbox\mktitle@bx\par\medskip
    \centering
    {\large\sffamily
      Zhiting Zhou\textsuperscript{1}, Xingchen Liu\textsuperscript{2}, Xinglin Yu\textsuperscript{3}, Jiashen Chen\textsuperscript{4}, Haoyang Wang\textsuperscript{1}, Jingao Xu\textsuperscript{5}, Yunhao Liu\textsuperscript{6}, and Xinlei Chen\textsuperscript{1}\par}%
    \vskip 0.45em
    {\small
      \begin{tabular}{@{}c@{}}
        \textsuperscript{1}Shenzhen International Graduate School, Tsinghua University;
        \textsuperscript{2}Sun Yat-sen University;
        \textsuperscript{3}Harbin Institute of Technology, Shenzhen\\
        \textsuperscript{4}Northwest University;
        \textsuperscript{5}The University of Hong Kong;
        \textsuperscript{6}Tsinghua University\\
        \{zhou-zt21,haoyang-22,yunhao\}@mails.tsinghua.edu.cn;
        chen.xinlei@sz.tsinghua.edu.cn\\
        liuxc7@mail2.sysu.edu.cn;
        220210522@stu.hit.edu.cn;
        2022115102@stumail.nwu.edu.cn;
        jingaoxu@hku.hk
      \end{tabular}\par}%
    \par\bigskip}}
\makeatother

\begin{document}

\title{EventRadar: Long-Range Visual UAV Discovery through Spatiotemporal Event Sensing}


\begin{abstract}
Unauthorized unmanned aerial vehicle (UAV) activity around airports, public venues, and other sensitive sites has made protected-airspace monitoring increasingly important. A practical sensing system must search a wide angular region, find small long-range targets, and return both bearing support and UAV-specific evidence before a restricted perimeter is breached. Existing UAV detection paths often rely on spatially organized evidence, such as body extent, silhouette, or track continuity. At long range, however, these cues become difficult to preserve and verify as the target footprint weakens and its image-plane support shrinks. \sysname\ follows a complementary cue: propeller-induced temporal periodicity, which recent event-camera sensing studies have shown can reveal UAV-specific motion after appearance becomes weak. We extend this cue to kilometer-scale active sensing with an event-camera prototype. Scene-Anchored Geometry Evidence (\sage) fuses scanning events with IMU pose to maintain a bearing-indexed scene memory, separating transient candidate support from persistent background clutter. Comb-guided Harmonic-Group Learned Iterative Shrinkage and Thresholding Algorithm (\chg) then treats each candidate as a weak high-rate timing signal and recovers phase-insensitive harmonic evidence with fixed compute. Compared with related event-camera baselines on 700--1500\,m UAV event recordings, \sysname\ achieves 0.990 mAP$_{.3}$ and 0.949 F1$_{.3}$, reduces FN$_{.3}$ to 0.009, and shows real-time feasibility in prototype profiling.
\end{abstract}



\maketitle
\thispagestyle{plain}
\pagestyle{plain}

\section{Introduction}

The global anti-UAV market is projected to grow from USD~4.48~billion in 2025 to USD~14.51~billion by 2030, reflecting rising demand to protect airports, energy facilities, public venues, and other sensitive sites from unauthorized UAV activity~\cite{marketsandmarkets_antidrone_2030,cisa_protect_critical_public_gatherings_2026,faa_restricting_drones_critical_infrastructure_2026}.
Safeguarding these environments requires pre-emptive detection of unauthorized UAV activity before a restricted perimeter is breached.
A practical protected-airspace sensing system must continuously search a wide angular region, identify small long-range UAVs, and return actionable bearing data together with UAV-specific evidence while suppressing false positives~\cite{cisa_uas_detection_technology_guidance}.

Existing UAV detection methods can be broadly viewed through two paths.
The traditional path seeks spatially organized evidence.
Across radar~\cite{radar_review,drone_detection_review_2024}, acoustic~\cite{uav_acoustic_bpf2025,AIM,acoustic_UAV}, thermal, and visual systems~\cite{rgb_review,rgb_detection,long_distance_rgb}, the evidence often appears as body extent, silhouette, local contrast, event density~\cite{evdetmav}, track continuity~\cite{ev_uav}, propeller-local structure~\cite{evpropnet,quad_prop_attr,mmUAVsense}, or other spatially organized cues.
Such cues are effective when the target leaves sufficient organized evidence in the sensing stream, but far-range protected-airspace sensing makes this assumption increasingly fragile.
As the UAV's physical footprint weakens and its image-plane support shrinks, spatially organized cues become difficult to preserve and verify. Table~\ref{tab:cue_behavior} gives a compact visual summary of these cue behaviors under long-range drone-detection stressors.

\begin{figure}[t]
  \centering
  \includegraphics[width=\linewidth]{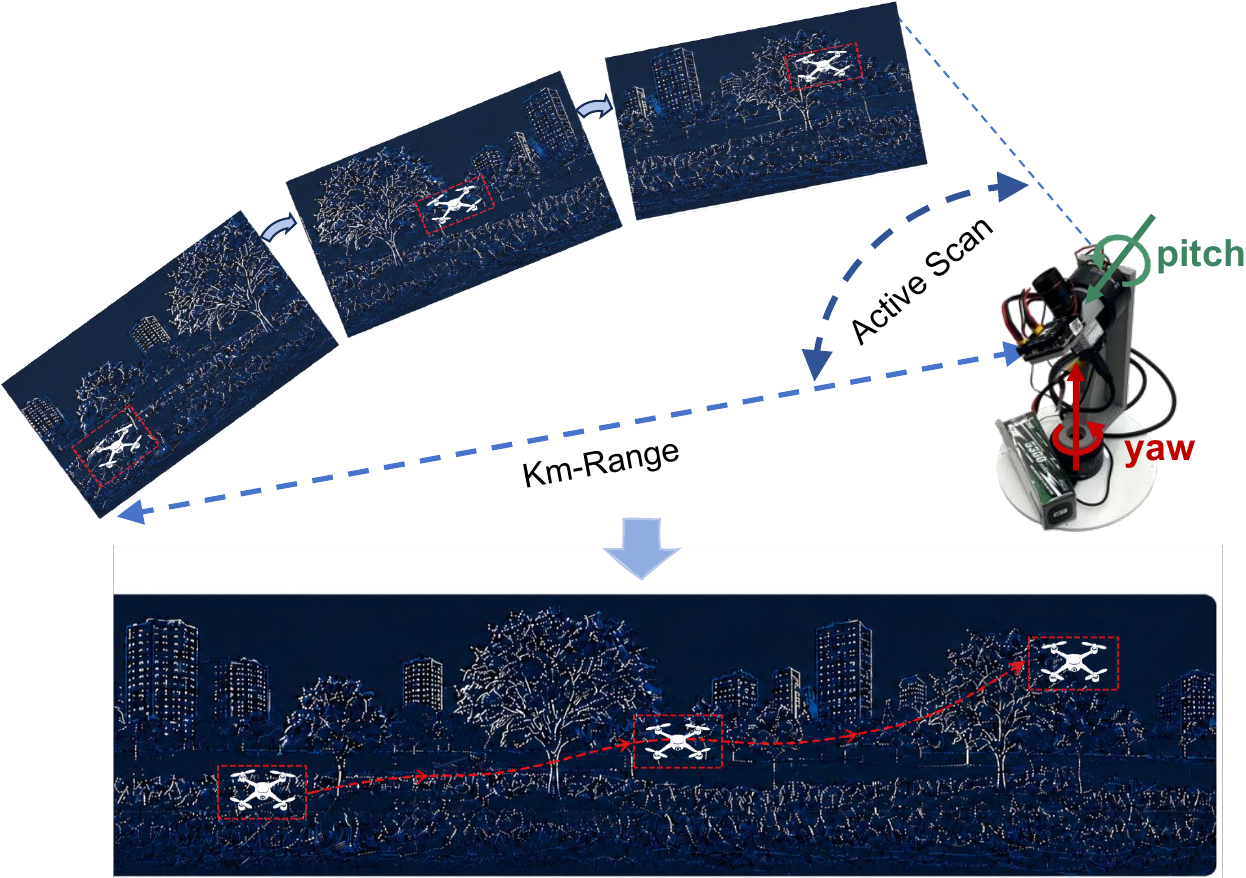}
  \caption{\sysname.}
  \Description{.}
  \label{fig:system_device}
\end{figure}

\begin{table*}[t]
  \centering
  \caption{Cue behavior under long-range drone-detection stressors.}
  \label{tab:cue_behavior}
  \setlength{\tabcolsep}{2.0pt}
\renewcommand{\arraystretch}{1.10}
\begin{tabular}{@{}>{\centering\arraybackslash}m{0.085\linewidth}*{5}{>{\centering\arraybackslash}m{0.173\linewidth}}@{}}
\toprule
 & \begin{minipage}[c]{\linewidth}\centering\normalsize \textbf{EvDetMAV}~\cite{evdetmav}\end{minipage} & \begin{minipage}[c]{\linewidth}\centering\normalsize \textbf{EVPropNet}~\cite{evpropnet}\end{minipage} & \begin{minipage}[c]{\linewidth}\centering\normalsize \textbf{effqpdet}~\cite{quad_prop_attr}\end{minipage} & \begin{minipage}[c]{\linewidth}\centering\normalsize \textbf{EV-SpSegNet}~\cite{ev_uav}\end{minipage} & \begin{minipage}[c]{\linewidth}\centering\normalsize \textbf{EventRadar\textsuperscript{*}}\end{minipage} \\
\midrule
\textbf{Cue}\textsuperscript{\dag} & \begin{minipage}[c]{\linewidth}\centering\small Layout\\[-1pt]\small event morphology\end{minipage} & \begin{minipage}[c]{\linewidth}\centering\small Shape prior\\[-1pt]\small propeller response\end{minipage} & \begin{minipage}[c]{\linewidth}\centering\small Rate burst\\[-1pt]\small event density\end{minipage} & \begin{minipage}[c]{\linewidth}\centering\small Point cloud\\[-1pt]\small frame stack + x-y-t tube\end{minipage} & \begin{minipage}[c]{\linewidth}\centering\small Harmonic\\[-1pt]\small ROI spectrum\end{minipage} \\
\addlinespace[5.5pt]
\textbf{700 m} & \includegraphics[height=0.80in,width=\linewidth,keepaspectratio]{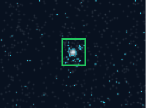} & \includegraphics[height=0.80in,width=\linewidth,keepaspectratio]{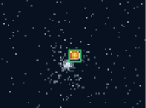} & \includegraphics[height=0.80in,width=\linewidth,keepaspectratio]{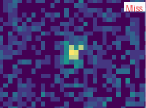} & \includegraphics[height=0.80in,width=\linewidth,keepaspectratio]{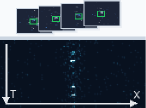} & \includegraphics[height=0.80in,width=\linewidth,keepaspectratio]{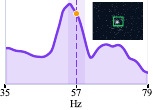} \\
\addlinespace[2.0pt]
\textbf{1500 m} & \includegraphics[height=0.80in,width=\linewidth,keepaspectratio]{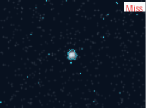} & \includegraphics[height=0.80in,width=\linewidth,keepaspectratio]{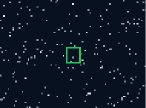} & \includegraphics[height=0.80in,width=\linewidth,keepaspectratio]{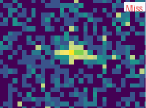} & \includegraphics[height=0.80in,width=\linewidth,keepaspectratio]{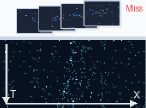} & \includegraphics[height=0.80in,width=\linewidth,keepaspectratio]{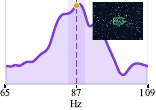} \\
\addlinespace[2.0pt]
\textbf{Cue behavior} & \begin{minipage}[c]{\linewidth}\centering\small layout vanishes at range\end{minipage} & \begin{minipage}[c]{\linewidth}\centering\small shape response drifts\end{minipage} & \begin{minipage}[c]{\linewidth}\centering\small rate cue is ambiguous\end{minipage} & \begin{minipage}[c]{\linewidth}\centering\small trajectory breaks in motion\end{minipage} & \begin{minipage}[c]{\linewidth}\centering\small harmonic peak remains\end{minipage} \\
\bottomrule
\end{tabular}
\par\vspace{2pt}
\begin{minipage}{0.985\linewidth}
\footnotesize\raggedright
\textsuperscript{\dag} For the four spatial baselines, long range removes the evidence each cue needs: layout/morphology loses body support, shape-prior response loses resolved blades, rate bursts mix with clutter/noise peaks, and point-cloud x-y-t tubes fragment under sparse motion.\\[-1pt]
\textsuperscript{*} EventRadar uses harmonic ROI spectra; this cue can remain visible because rotor frequency is set by propeller actuation rather than projected size.
\end{minipage}

\end{table*}

\begin{figure*}[t]
  \centering
  \includegraphics[width=\textwidth]{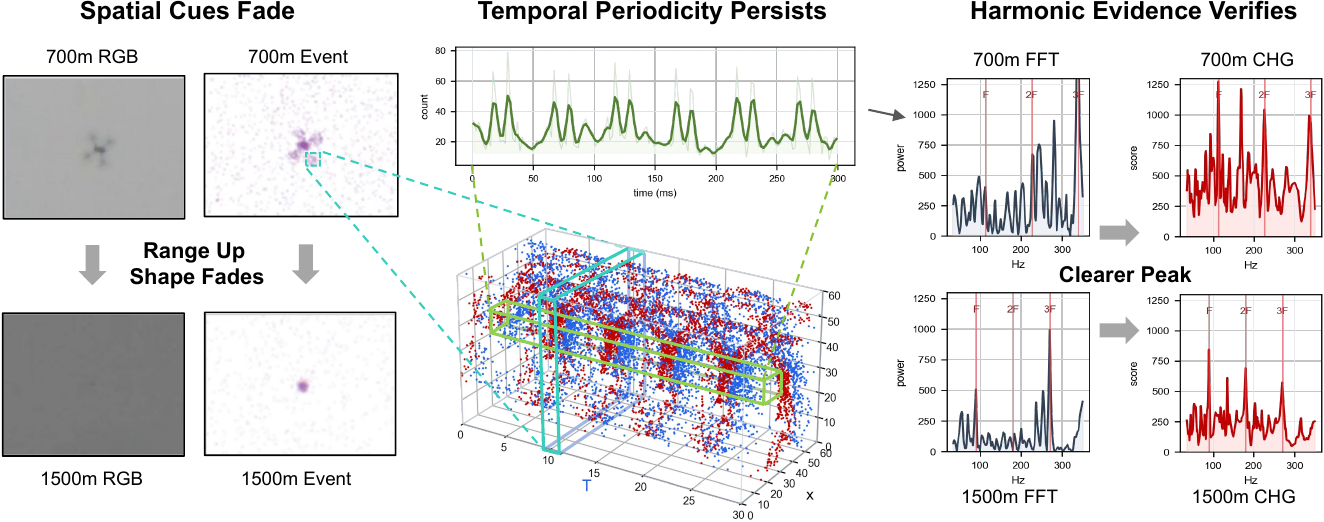}
  \caption{Long-range event-sensing observation.}
  \Description{Three-panel observation figure: spatial cues fade with range, temporal periodicity persists in event timing, and harmonic evidence verifies the candidate through spectral and CHG-LISTA representations.}
  \label{fig:motivation}
\end{figure*}

A more recent path exploits temporal evidence induced by propeller rotation.
The continuous rotation of UAV propellers generates stable, high-frequency periodic micro-motion~\cite{uav_acoustic_bpf2025,evpropnet,event_propeller_sensing}.
This dynamic signature is difficult to exploit with conventional frame cameras, whose frame-rate and spatial-sampling limits keep recognition tied to appearance cues~\cite{rgb_review,long_distance_rgb}.
Event cameras, with asynchronous microsecond-level temporal resolution, provide a high-rate interface for observing such periodic signals, and recent work has shown their potential for UAV identity verification, and internal flight-state inference~\cite{mmeloc,count_every_rotation}.
These studies show that temporal propeller sensing has become a promising evidence channel for UAV perception, with figure~\ref{fig:motivation} illustrates this cue shift.
We therefore ask: \textit{Can temporal propeller evidence extend the effective identification distance of visual UAV detection systems?}

\noindent \textbf{Our work.}
To answer this question, we propose \textbf{EventRadar}, a long-range UAV detection system driven by periodic micro-motion of UAV propellers.
EventRadar actively sweeps distant airspace to identify candidate bearings, subsequently isolating and verifying targets within these proposed regions, as shown in Figure \ref{fig:system_device}.
Rather than attempting to recover complete spatial appearance from just a few pixels, EventRadar treats each candidate direction as a continuous, high-rate temporal observation and verifies whether it carries distinct propeller-periodic evidence.

However, making EventRadar a practical system in the wild and elevating the effective identification distance of visual UAV detection systems face two challenges.

\noindent $\bullet$ \textbf{C1:}
\textit{Spatially, where to see: scanning motion-induced environment events overwhelm the faint UAV event stream, obscuring candidate bearings.}
Event cameras are highly sensitive to dynamic brightness changes.
When the sensor actively scans expansive airspace, camera motion triggers dense background events across the image plane.
As a result, weak UAV-triggered signals can be overwhelmed by environmental clutter before the system can isolate a candidate bearing, making far-range protected-airspace sensing a candidate discovery problem, rather than merely a local object detection task.

\noindent $\bullet$ \textbf{C2:}
\textit{Temporally, how to see: UAV periodic micro-motions are transient and weak across only a few pixels, hindering reliable long-range recognition.}
Once a candidate region is proposed, the system must verify it from a short and noisy event window.
Because event generation depends on dynamic contrast, the captured signal rarely resembles a clean sinusoid.
Instead, the temporal sequence can contain spatial drift, missing actuation cycles, and non-target transients.
Recovering reliable rotor-periodic evidence from such fragmented event timing is therefore central to long-range UAV identification.

EventRadar is organized around two core designs.

\noindent $\bullet$ To solve \textbf{C1}, we propose the Scene-Anchored Geometry Evidence module (\textbf{\sage}), driven by the insight that active visual sensing makes events move rapidly on the image plane, while the underlying background structure remains more stable in a world-centric directional space.
\sage\ tightly fuses the asynchronous event stream with real-time IMU and gimbal pose data to construct and maintain a bearing-indexed scene memory.
In this representation, scanning events are accumulated as stable scene evidence rather than isolated image-plane detections, allowing the system to identify candidate bearings under wide-area active sensing.

\noindent $\bullet$ To solve \textbf{C2}, we propose the Comb-guided Harmonic-Group LISTA module (\textbf{\chg}) for temporal verification.
Rather than treating a microscopic candidate as a static image patch, \chg\ treats its asynchronous events as a weak, high-rate temporal observation.
It acts as a fixed-budget harmonic-evidence recovery module that converts each candidate Region of Interest (ROI) into an event signal, removes slow drift, pools phase-insensitive harmonic groups, and scores local spectral ratios as rotor-frequency evidence.
By directly scoring harmonic structure, \chg\ translates fragmented temporal cues into candidate-level verification evidence for final UAV recognition.

In summary, this paper makes the following contributions.\\
\textit{(i)} We extend temporal propeller sensing to kilometer-scale UAV discovery, showing that propeller-periodic event evidence can remain useful after spatial appearance becomes unreliable at distance.\\
\textit{(ii)} We propose EventRadar, an active rotating event-camera sensing prototype, combining \sage\ for scene-anchored candidate discovery and \chg\ for candidate-level temporal verification.\\
\textit{(iii)} We evaluate EventRadar on long-range UAV detection tasks from 700\,m to 1.5\,km and compare it with related event-camera baselines, demonstrating stronger detection availability, missed-target control, and prototype real-time feasibility.

\section{Background and Related Work}
\label{sec:background}

\subsection{Anti-UAV Sensing and Long-Range Evidence}
\label{sec:problem_definition}

Anti-UAV systems differ in the evidence they expose to a response loop. Radar and mmWave sensing can estimate range, velocity, angle, radar cross section, tracks, and micro-motion signatures~\cite{drone_detection_review_2024,radar_review,UAV_bird,mmUAVsense}. Radio Frequency (RF) systems monitor control, telemetry, video, Wi-Fi, Bluetooth, or other emissions, while acoustic systems use motor and rotor spectra together with array-based direction cues~\cite{rf_sdr_survey,AIM,acoustic_UAV,uav_acoustic_bpf2025}. These channels can support detection under favorable conditions, but long-range protected-airspace monitoring stresses their discriminative evidence: radar cross section, clutter, multipath, radio silence or interference, outdoor attenuation, wind, and array geometry can all make a small UAV difficult to verify.

This paper focuses on the evidence interface provided by event-based optical sensing. A visual response loop needs candidate bearing and image-plane support, but at long range a UAV may occupy only a few pixels. The central question is not whether non-visual sensing is useful, but how an optical or event system can still provide UAV-specific evidence when spatial appearance becomes weak.

\subsection{Visual and Event-Based Small-Object Detection}
\label{sec:cues_degradation}

Visual small-object detectors localize targets from spatial cues such as local contrast, edges, texture, silhouette, motion consistency, or learned appearance features~\cite{rgb_detection,rgb_review,long_distance_rgb}. These cues support near-range localization, but their discriminative value decreases as distance reduces the target to tiny image support. The detector may still find a suspicious point or blob, while the evidence needed to verify that it is a UAV has already degraded.

Event cameras change the sensing principle by reporting asynchronous brightness-threshold crossings with high temporal resolution, but many event-camera detectors still choose spatial or motion-density representations. Existing methods often convert events into frames, voxels, tensors, point-cloud-like features, density maps, local motion cues, recurrent features, or trajectory consistency before detection or segmentation~\cite{gen1_event_detection,rvt_event_detection,ev_uav,ev_flying,m2e_uav,drone_detection_event_cameras}. Such cues are useful for candidate discovery, yet they remain brittle for far-range UAV verification: compact density peaks and short tracks can also arise from swept background edges, moving clouds, sensor noise, vibration, birds, or other small objects~\cite{rvt_event_detection,ev_flying,m2e_uav}.

\subsection{Frequency Signatures for UAV Verification}
\label{sec:propeller_frequency_cue}

Frequency-domain signatures provide a complementary way to read target evidence because they describe periodic physical processes rather than spatial appearance. Radar and mmWave systems use micro-Doppler or blade-induced rotary structure to separate UAVs from other objects~\cite{UAV_bird,radar_review,mmUAVsense}. Acoustic sensing observes rotor tones, blade-pass frequencies, and harmonic structure~\cite{AIM,acoustic_UAV,uav_acoustic_bpf2025}. Event cameras have also been used to estimate flicker or repetitive visual signals~\cite{frequency_cam,event_periodic_signal,s_rope,fries_event_surveillance}, and propeller-focused event-camera works show that rotor motion can generate visual timing evidence in controlled or shorter-range settings~\cite{evpropnet,virtual_fence,quad_prop_attr,event_propeller_sensing,ddhf}.

These works show that periodic evidence can be discriminative, but long-range active event sensing introduces a coupled problem. The system must first find tiny candidate support inside dense scan-induced background events, then recover reliable harmonic evidence from a short and noisy event window. Existing methods address parts of this pipeline, but they do not provide a unified path from background-aware long-range candidate discovery to candidate-level propeller-frequency verification. This gap motivates \sysname's separation between \sage\ spatial evidence organization and \chg\ temporal harmonic evidence recovery.

\section{System Overview}
\label{sec:overview}

\begin{figure}[t]
  \centering
  \includegraphics[width=\linewidth]{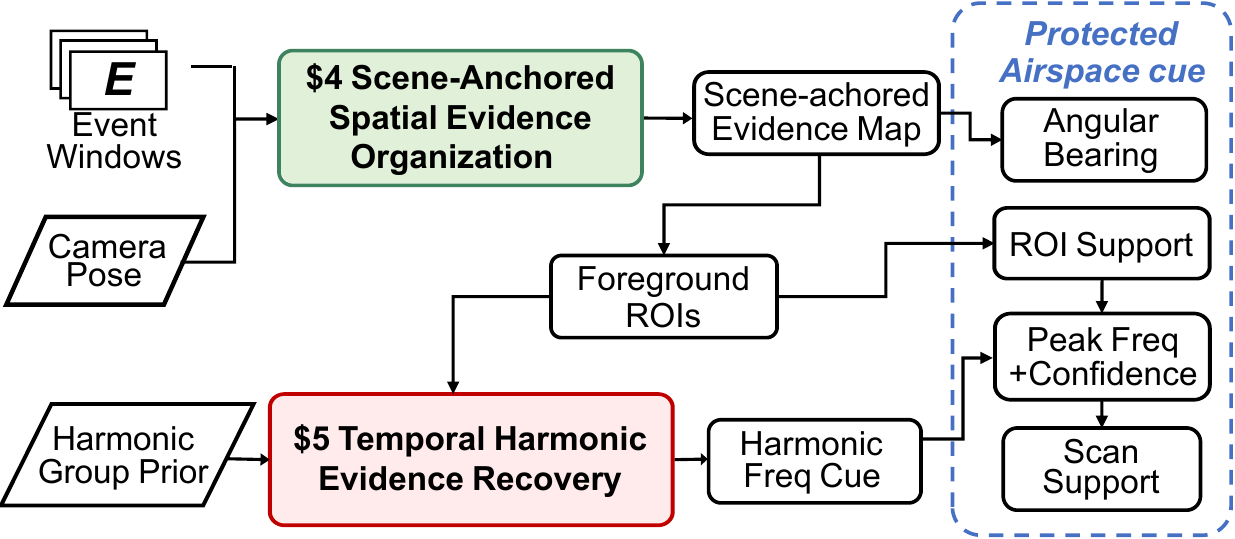}
  \caption{Overview of \sysname.}
  \Description{System overview figure showing EventRadar's protected-volume scan, scene-anchored spatial evidence organization with a bearing-indexed scene representation, CHG-LISTA harmonic evidence recovery, and protected-airspace bearing, bbox, frequency-confidence, and scan-support outputs.}
  \label{fig:system_overview}
\end{figure}

Figure~\ref{fig:system_overview} shows how \sysname\ turns the frequency-cue insight into a practical sensing framework. The input is a stream of scanning events together with time-aligned camera attitude from the gimbal and IMU. The spatial stage, \textit{\sage}, organizes event evidence in a scene-anchored, bearing-indexed representation. This representation gives candidate discovery scene context: transient local evidence, distinguished from persistent background, can be proposed for temporal harmonic recovery.

The temporal stage receives candidate regions, their bearing context, and their local event support. Each candidate ROI is converted into a short high-rate event-count signal and passed to \textit{\chg} to report propeller-frequency confidence for each candidate. Thus, \sysname\ uses \textit{\sage} to discover candidate regions and \textit{\chg} to test whether those regions carry UAV-specific propeller cues. The output of \sysname\ is a protected-airspace cue: candidate bearing, bbox support, frequency estimate, and confidence can guide the next scan decision. 

\section{Spatial Evidence Organization for Candidate Discovery}
\label{sec:candidate_selection}

\subsection{Design Insight and Workflow}
\label{sec:active_candidate_selection}

At long range, a moving UAV often appears not as a recognizable object, but as a small event cluster sustained by motion. Morphology, texture, and silhouette may be too weak to discriminate the target, so candidate discovery must first identify spatiotemporal event support whose continuity makes it worth temporal verification.

Active scanning complicates this cue because static background structures are deliberately swept across the sensor and can form compact short-window peaks. The useful distinction appears in a scene-anchored coordinate frame: background events tend to recur at stable world directions, whereas a moving target traces neighboring directions and is transient at each one. \sage\ organizes scanning events in a scene-anchored direction space, preserving image-plane ROI support while using a bearing-indexed angular memory to separate persistent background context from transient, spatially continuous candidates for harmonic verification. This spatial layer outputs candidate support, bearing estimates, and scene context, rather than making a final decision by only geometry.

The module proceeds in three steps. It first maps image-plane events into scene-anchored directional evidence using camera pose (Section~\ref{sec:event_projection_pose}), then accumulates this evidence as an angular scene memory that captures persistent background context (Section~\ref{sec:polar_scene_evidence}), and finally combines local event support with this scene context to propose candidate regions, bearings, and bbox support for temporal harmonic verification (Section~\ref{sec:background_aware_candidate}).

\subsection{From Image Events to Directional Evidence}
\label{sec:event_projection_pose}

For an event or candidate pixel $\mathbf{u}=(u,v,1)^\top$, \sysname\ first maps the pixel to a camera-frame unit ray using the camera calibration and undistortion model:
\begin{equation}
  \mathbf{p}^c =
  \frac{K^{-1} f^{-1}(\mathbf{u})}
  {\|K^{-1} f^{-1}(\mathbf{u})\|_2}.
\end{equation}
With a time-aligned camera-to-world rotation $R_t$ from gimbal feedback in our prototype, or from IMU attitude and sensor fusion in other mountings, the scene-anchored viewing ray is
\begin{equation}
  \mathbf{p}^s = R_t\mathbf{p}^c .
\end{equation}
When the base is fixed, $R_t$ can be constructed from the measured yaw and pitch of the gimbal plus the camera-to-gimbal calibration. With a rigidly mounted IMU, the rotation can instead be obtained from the interpolated camera attitude. In either case, the pose stream defines a common polar direction space in which image-plane events from different scan poses can be compared.

\begin{figure}[t]
  \centering
  \includegraphics[width=\linewidth]{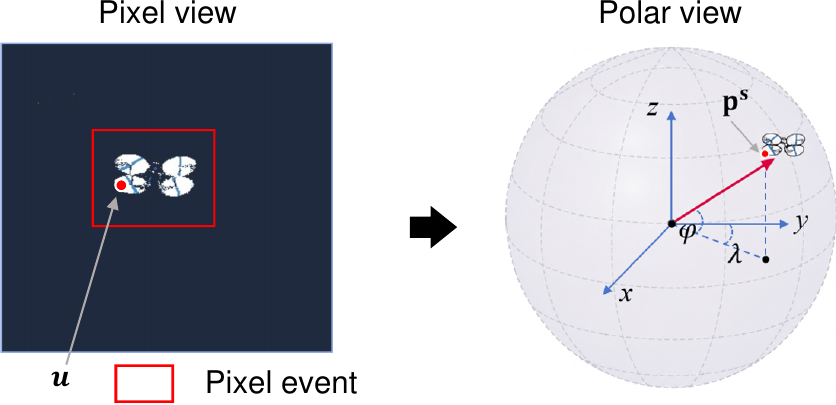}
  \caption{Scene-anchored coordinate geometry.}
  \Description{A polar coordinate diagram showing a global unit viewing ray, azimuth and elevation angles, a bearing point, and formulas mapping a pixel event to polar angular coordinates.}
  \label{fig:polar_projection}
\end{figure}

Figure~\ref{fig:polar_projection} illustrates the coordinate geometry used by the spatial layer. The point is not the coordinate conversion itself, but the physical meaning of the two coordinate systems. The image coordinate $\mathbf{u}$ locates event evidence on the two-dimensional image plane, while the spatial vector $\mathbf{p}^s$ locates the viewed direction in the scene-frame coordinate system. During active scanning, background-induced events may move to different image locations, but projection through the current pose maps them to similar spatial vectors. A tiny moving UAV, in contrast, may appear only as a weak local image cluster and remain difficult to identify from spatial appearance alone; its usefulness comes from the time-local, motion-induced support introduced in Section~\ref{sec:active_candidate_selection}. The projection step therefore supplies the common coordinate frame needed to compare persistent scene directions with transient moving support.

\subsection{Angular Scene Evidence Accumulation}
\label{sec:polar_scene_evidence}

Once events are expressed as $\mathbf{p}^s=(p_x^s,p_y^s,p_z^s)$, they are indexed by azimuth and elevation:
\begin{equation}
  \lambda = \mathrm{atan2}(p_x^s,p_z^s),
  \qquad
  \varphi = \arcsin(-p_y^s).
\end{equation}
The accumulated representation stores polar angular scene evidence as a bearing-indexed memory. Persistent structures, such as horizon lines, buildings, and static texture, reappear in nearby angular bins during repeated scans. A small UAV candidate, by contrast, is expected to form a local vector cluster that moves through neighboring angular bins, so its support is spatially continuous but not persistent at one fixed scene direction.

This memory is defined by three operations. First, calibration turns each event pixel into an undistorted camera-frame unit ray, so nearby pixels represent nearby viewing directions rather than arbitrary image samples. Second, the time-aligned scan pose, supplied in our prototype by the gimbal angle stream, rotates those rays into a common scene frame. Third, the resulting scene-frame vectors are accumulated as a regional-density-managed unit-sphere event map. The underlying memory stores scene-frame unit rays, while latitude--longitude cells regulate angular density and provide the polar readout used for visualization and candidate context. Event counts, recency, and local temporal statistics can then be attached to this directional support. In our prototype realization, events are undistorted into unit rays, accumulated on a sphere under estimated rotations, maintained as a global event-sphere map, and rasterized into polar panoramic event maps. We use this representation as scene context for proposal, while treating candidate stability and false-candidate suppression as evaluation targets rather than assumed consequences of the projection step.

The temporal statistic used for this context is a persistence ratio over a recent scan window: for each angular cell, we measure the fraction of short time bins in which that cell keeps producing events. This makes the temporal cue visible without adding a separate time axis. A persistent background direction, hot pixel pattern, or repeatedly swept static edge remains active across many bins and is therefore assigned high persistence. A moving UAV activates neighboring directions along a continuous path, but it stays in any single direction only briefly; its response is spatially continuous yet temporally transient. In the transient--persistent visualization, red denotes low-persistence transient support and blue denotes high-persistence background context.

\begin{figure}[t]
    \centering
    \includegraphics[width=\linewidth]{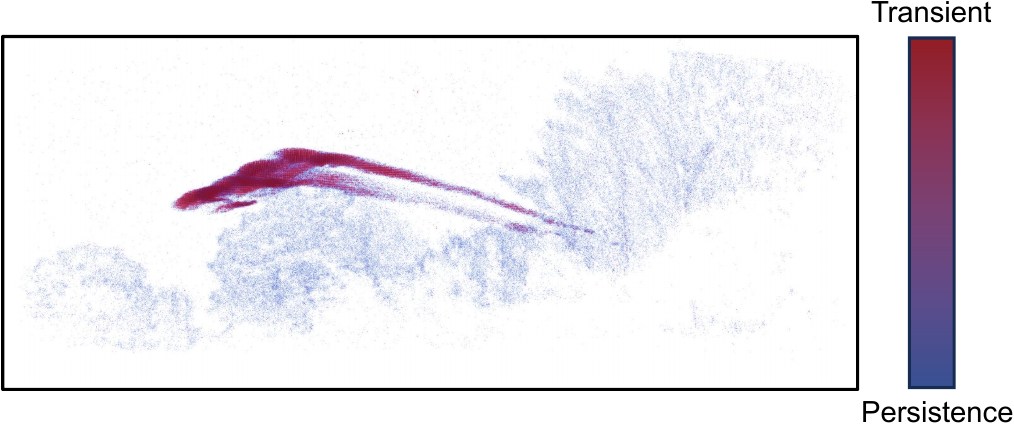}
    \caption{Transient--persistent evidence for ROI proposal.}
    \Description{A two-panel figure. The first panel shows local scan windows, a polar scene memory with persistent background regions and a transient candidate bin, a background-aware proposal gate, and candidate outputs passed to harmonic recovery. The second panel shows a red spatially continuous target trail over blue persistent background activity.}
    \label{fig:polar_scene_memory}
\end{figure}

\subsection{Background-Aware Candidate Proposal}
\label{sec:background_aware_candidate}

Candidate proposal combines local event evidence with the scene-anchored polar representation. Figure~\ref{fig:polar_scene_memory}(a) shows how local image-plane support is interpreted through the polar scene memory, so proposal is conditioned on both recent local activity and scene-persistent context. A local event cluster is considered a candidate when, in the polar representation, it carries enough recent support to form a time signal and exhibits the spatiotemporal pattern expected from a moving object: temporal transience together with spatial motion relative to persistent background context. In Figure~\ref{fig:polar_scene_memory}(b), this appears as a red trajectory-like support pattern separated from the blue persistent scene response. This proposal layer can use adaptive event clusters and event-count peaks as local evidence, but the bearing-indexed memory provides the background-aware context needed for active scanning.

Figure~\ref{fig:polar_scene_memory} summarizes the proposal interface. \sage\ does not need to decide whether the object is a UAV. Instead, it identifies candidate support that deserves temporal harmonic verification. The geometry-aware layer preserves the image ROI needed to extract events, the polar bearing needed for scan context, and the local support needed to rank candidate readouts. The output is a set of candidate regions
\begin{equation}
  \mathcal{C}_t = \{(\Omega_m,\lambda_m,\varphi_m,b_m,t_m)\}_{m=1}^{M_t},
\end{equation}
where $\Omega_m$ is the image or event-window ROI used for local temporal extraction, $(\lambda_m,\varphi_m)$ is its polar bearing, $b_m$ is the local bbox support, and $t_m$ is the event-window timestamp. The proposal score should be interpreted as candidate support, not as a final UAV decision.

\sage\ produces candidate regions, bearing estimates, and scene context that define the input contract for temporal harmonic recovery. Given this candidate support, \chg\ recovers propeller-frequency evidence inside each region as described in Section~\ref{sec:chg_lista}. Bearing memory and scan scheduling can build on the same scene-anchored representation, while the central role here is to provide stable spatial evidence for candidate discovery under wide-area active sensing.

\section{Temporal Harmonic Evidence Recovery for Candidate Verification}
\label{sec:chg_lista}

\subsection{Design Insight and Workflow}
\label{sec:timeseries}

\begin{figure}[t]
  \centering
  \includegraphics[width=\linewidth]{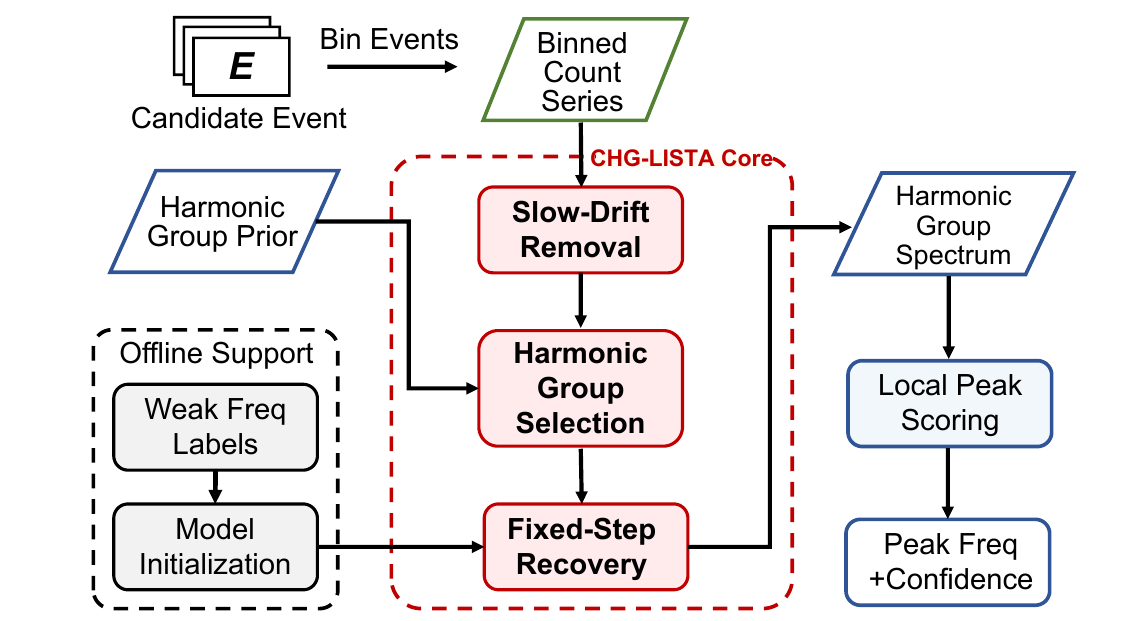}
  \caption{\chg\ workflow.}
  \Description{A compact CHG-LISTA workflow diagram showing candidate events converted into a binned count series, followed by slow-drift removal, harmonic group selection, fixed-step recovery, local peak scoring, and peak frequency plus confidence output. An offline weak-label path supports training and auditing.}
  \label{fig:chg_pipeline}
\end{figure}

\sage\ narrows the search to candidate support, but the events inside that support are not a clean rotor waveform. At long range, the signal is often a short, low-count pulse train with missing cycles, arbitrary phase, and slow drift from scan motion or background activity. A direct appearance verifier has too little spatial structure to read, while a raw Fast Fourier Transform (FFT) peak is brittle because propeller evidence spreads across harmonics and competes with a changing local spectral floor.

The design insight behind \chg\ is to verify a candidate by recovering a compact harmonic explanation rather than by detecting a single spectral peak. \chg\ first forms the candidate event signal (Section~\ref{sec:signal_formation}), prepares a nuisance-rejected harmonic representation (Section~\ref{sec:problem}), recovers group-level evidence with fixed compute (Section~\ref{sec:lista}), and converts the recovered spectrum into candidate-level frequency and confidence evidence (Section~\ref{sec:score_output}). The weak frequency-supervision path in Figure~\ref{fig:chg_pipeline} supports training and auditing, and is described with the implementation details in Section~\ref{sec:algorithm_impl}.

\subsection{Temporal Signal Formation from Candidate Events}
\label{sec:signal_formation}

\subsubsection{Candidate event binning.}
Given a candidate support $\Omega$ from \sage\ and an event window of length $T$, the temporal stage uses the events inside that support as its measurement. This readout is tied to an existing candidate and is not a separate semantic detection result. Each event is $e_i=(x_i,y_i,t_i,p_i)$, where $(x_i,y_i)$ is the pixel location, $t_i$ is the timestamp, and $p_i$ is the polarity. \sysname\ forms an $N$-sample count series by binning the events in $\Omega$ over the window:
\begin{equation}
  s_\Omega[n] =
  \sum_i \mathbf{1}\{(x_i,y_i)\in\Omega,\; t_i-t_0 \in [nT/N,(n+1)T/N)\}.
  \label{eq:event_count}
\end{equation}
The online verifier uses this event-count signal as the input to harmonic recovery. Polarity is not used in the online forward pass, only in the weak-supervision procedure described in Section~\ref{sec:algorithm_impl}.

\begin{figure}[t]
    \centering
    \begin{subfigure}[t]{0.4\linewidth}
        \centering
        \includegraphics[width=\linewidth]{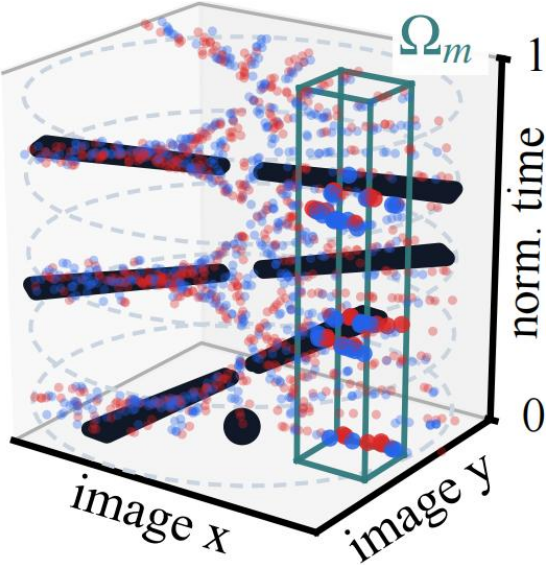}
        \caption{local rotor event volume}
        \label{fig:local_rotor}
    \end{subfigure}
    \hfill
    \begin{subfigure}[t]{0.4\linewidth}
        \centering
        \includegraphics[width=\linewidth]{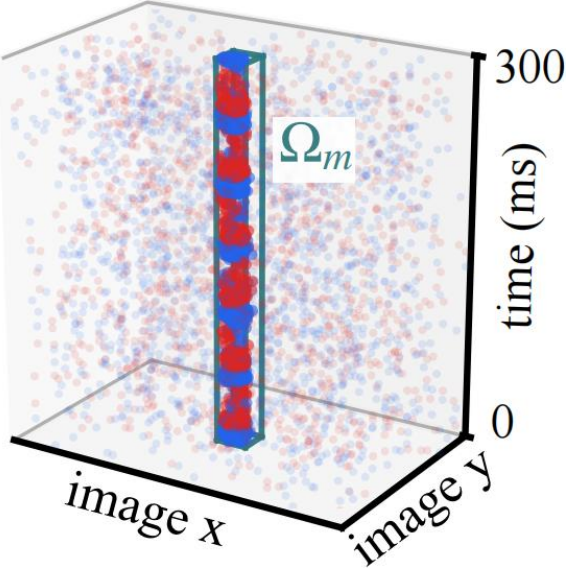}
        \caption{far-range candidate}
        \label{fig:far_range}
    \end{subfigure}
    \begin{subfigure}[t]{\linewidth}
        \centering
        \includegraphics[width=\linewidth]{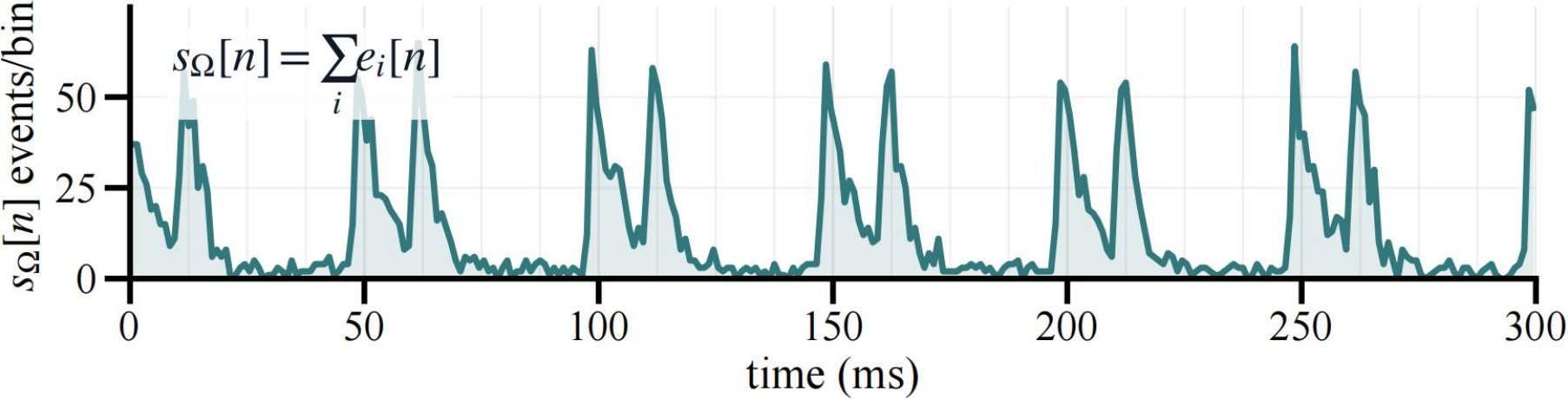}
        \caption{summed event count}
        \label{fig:summed_event}
    \end{subfigure}
    \begin{subfigure}[t]{\linewidth}
        \centering
        \includegraphics[width=\linewidth]{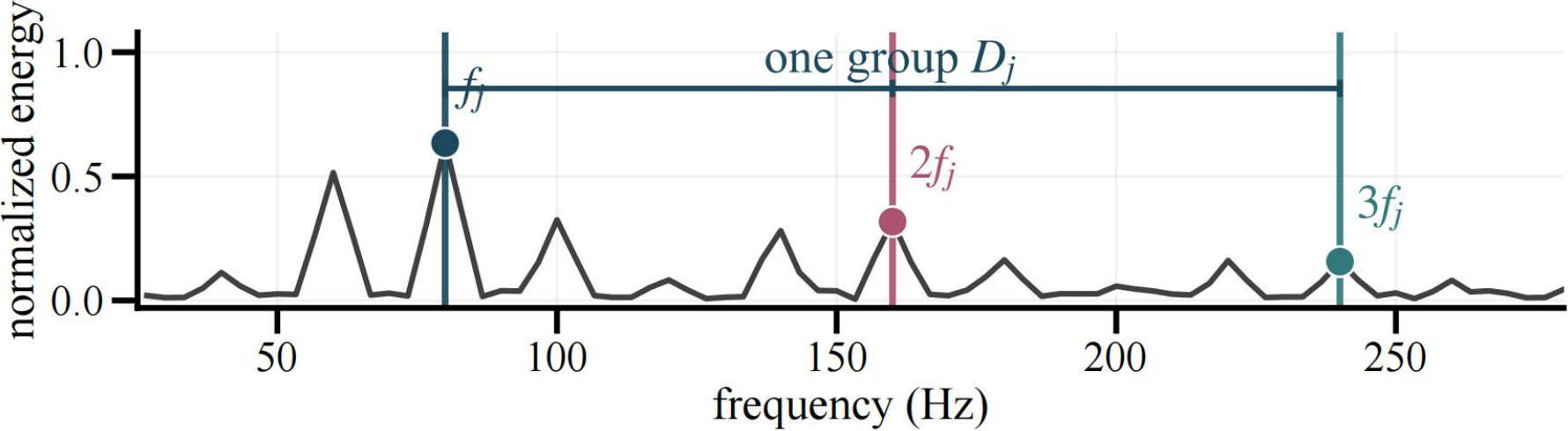}
        \caption{harmonic-group evidence}
        \label{fig:harmonic}
    \end{subfigure}
    
    \caption{Candidate event formation and temporal readout.}
    \Description{A four-panel single-column figure showing local rotor event-volume geometry, far-range candidate support pooling, a summed event-count signal, and harmonic-group evidence across the first three harmonics.}
    \label{fig:candidate_event}
\end{figure}

Figure~\ref{fig:candidate_event} illustrates this readout. Panel (a) shows the local rotor-induced event volume, panel (b) shows the far-range candidate support that pools the remaining activity, panel (c) shows the summed count signal defined in Eq.~\ref{eq:event_count}, and panel (d) shows why this signal is later interpreted as harmonic-group evidence rather than as a single spectral peak.

\begin{figure*}[t]
  \centering
  \includegraphics[width=\textwidth]{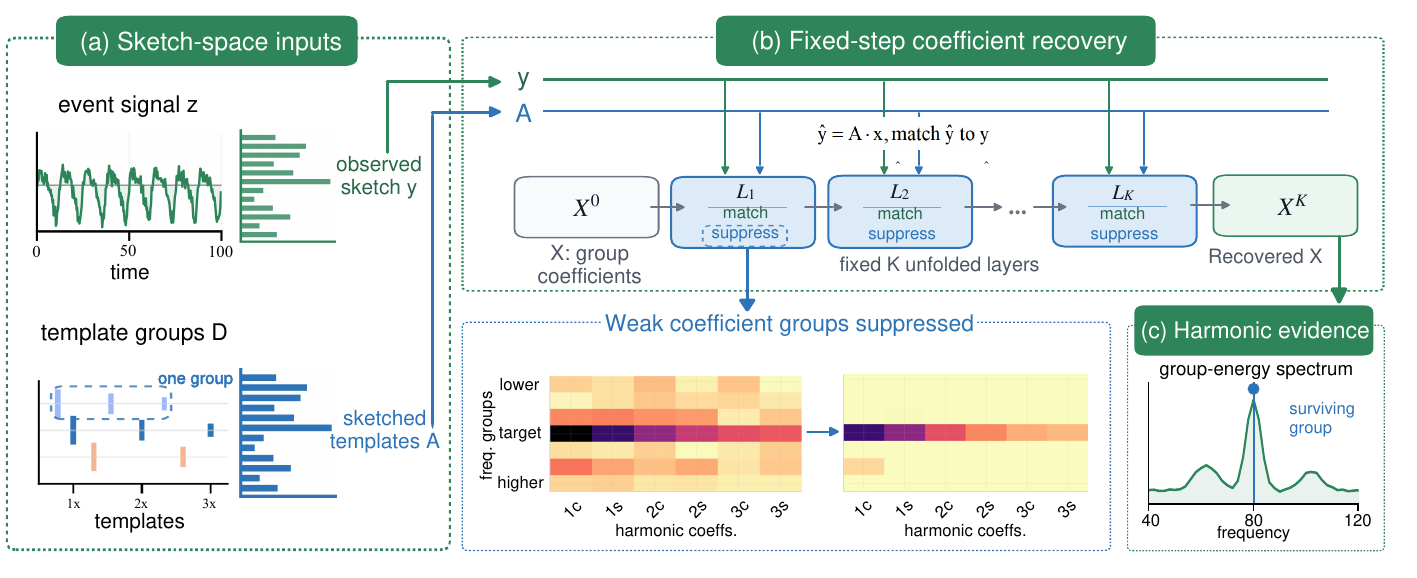}
  \caption{Fixed-step \chg\ recovery scheme.}
  \Description{A three-panel scheme showing sketch-space inputs from an event signal and template groups, fixed-step coefficient recovery with observed sketch and template-bank inputs through unfolded match-and-suppress layers, an inset comparing coefficient groups before and after suppression, and the final harmonic evidence spectrum.}
  \label{fig:chg_recovery_scheme}
\end{figure*}

\subsubsection{Verifier output contract.}
Temporal evidence formation outputs $\mathbf{s}\in\R^N$, a candidate-level time series. This representation intentionally discards most within-ROI spatial detail and keeps only the timing evidence needed for harmonic recovery. The following modules estimate rotor frequency and confidence from this signal, so the boundary with \sage\ remains clean: \sage\ supplies the candidate support, while \chg\ supplies harmonic evidence for that candidate.

\subsection{Harmonic Representation Preparation}
\label{sec:problem}

The count series from Section~\ref{sec:signal_formation} is next converted into a representation suitable for sparse harmonic recovery. The conversion has two goals. First, propeller evidence should be represented as a harmonic group, because an event-camera blade signal is pulse-like and rarely concentrates at one isolated sinusoid. Second, slow count variation from body drift, gimbal sweep, illumination changes, and static background edges should be removed before the solver searches for rotor periodicity. \chg\ therefore forms a signed, nuisance-rejected residual and represents that residual with phase-insensitive harmonic groups.

\subsubsection{Propeller harmonic-group prior.}
A multirotor UAV generates repeated brightness changes around blade edges as the propellers rotate. In an event camera, these changes appear as threshold-crossing pulses rather than as a clean sinusoid. The useful frequency cue is therefore not a single isolated spectral line. A weak propeller signal can spread across the rotor frequency, blade-pass frequency, and higher harmonics, with the observed envelope depending on blade visibility, contrast, duty cycle, and the event-generation threshold~\cite{uav_acoustic_bpf2025,evpropnet,event_propeller_sensing}.

This observation leads to the harmonic-group prior used by \chg. Instead of assigning each frequency bin an independent atom, \chg\ treats the harmonics that share one candidate base frequency as one structured unit.
The group collects evidence at $f_q,2f_q,3f_q,\ldots$ while preserving a single rotor-frequency identity. 
This design is crucial for far-range candidates, as sparse events and missing cycles can weaken individual harmonics even when the overall group reflects propeller motion.

After removing slow nuisance motion, the candidate signal is modeled as a few harmonic groups:
\begin{equation}
  z(t) =
  \sum_{q=1}^{Q}
  \sum_{h\in\mathcal{H}}
  a_{q,h}\cos(2\pi h f_q t+\phi_{q,h})
  +\eta(t),
  \label{eq:prop_signal}
\end{equation}
where $Q$ is small, $\mathcal{H}$ is the harmonic set, $f_q$ is a candidate rotor $1\times$ frequency, and $\eta(t)$ includes event noise and residual background motion. Under this model, a positive frequency cue should be explainable as a compact propeller-frequency group, not merely as a high event count.

\subsubsection{Slow-drift removal.}
\label{sec:preprocess}
Before building harmonic groups, \chg\ removes slow nuisance variation from the candidate time series. Let $\mathbf{s}\in\R^N$ be the binned signal and let
\begin{equation}
  P = [\mathbf{1},\mathbf{t},\mathbf{t}^2,\ldots,\mathbf{t}^q]\in\R^{N\times(q+1)}
\end{equation}
be the nuisance subspace spanned by low-order time polynomials. The projection
\begin{equation}
  P_\perp = I - P(P^\top P)^\dagger P^\top
\end{equation}
removes this subspace, producing $\tilde{\mathbf{s}}=\mathbf{s}P_\perp^\top$. 
Unless otherwise stated, \sysname\ uses a first-order polynomial nuisance model.

The projected signal is normalized with a median and median absolute deviation:
\begin{equation}
  \mathbf{z} =
  \frac{\tilde{\mathbf{s}}-\operatorname{median}(\tilde{\mathbf{s}})}
       {1.4826\,\operatorname{MAD}(\tilde{\mathbf{s}})+\epsilon}.
\end{equation}
The residual remains signed. This choice preserves contrast-change and phase information for the following cosine and sine dictionary, whereas rectification would discard both.

\subsubsection{Phase-insensitive harmonic groups.}
\label{sec:dictionary}
The harmonic model must also absorb unknown initial phase. \chg\ therefore builds one real-valued quadrature comb group for each candidate base frequency $f_j$:
\begin{equation}
  D_j(t)=
  \left\{
  w_h\cos(2\pi h f_j t),\;
  w_h\sin(2\pi h f_j t)
  \right\}_{h\in\mathcal{H}},
\end{equation}
where $\mathcal{H}=\{1,2,3\}$ by default and $w_h=1/h$ is a simple attenuation heuristic that prevents higher harmonics from dominating the group score. The cosine/sine pair is equivalent to a two-channel complex representation and absorbs unknown initial phase without requiring complex-valued deployment kernels. The group therefore acts like a comb template: it pools energy across $f_j,2f_j,3f_j,\ldots$ while preserving a single base-frequency identity.

Stacking all groups yields a dictionary $D\in\R^{N\times(2|\mathcal{H}|F)}$ for $F$ candidate frequencies. Each atom is normalized. Accordingly, $D$ is initialized with the event-domain harmonic-comb basis and refined during training, preserving the structured harmonic prior while accommodating sensor-specific pulse shapes, contrast responses, and lens-dependent artifacts.

\subsection{Fixed-Step Harmonic Recovery}
\label{sec:lista}

After the harmonic representation is built, \chg\ recovers frequency evidence with a fixed computation budget. Direct sparse solvers are a natural fit for this structure, but their support selection and repeated least-squares updates make online latency data-dependent. \chg\ therefore keeps the group-sparse recovery objective and unfolds a small, fixed number of solver steps into trainable layers. Figure~\ref{fig:chg_recovery_scheme} summarizes this fixed-budget path. The detrended candidate signal and harmonic dictionary are mapped into a compact sketch pair $(\mathbf{y},A)$; the unfolded layers then alternate residual matching and group-wise suppression to produce the coefficient tensor $X^K$ and its group-energy spectrum. In this design, recovery is performed over frequency-indexed harmonic groups, not isolated spectral coefficients, and the compute budget is fixed by construction.

\subsubsection{Group-sparse recovery objective.}
\label{sec:objective}
Let $\mathbf{z}\in\R^N$ denote the detrended and normalized candidate signal. A sparse binary sketch matrix $\Phi\in\R^{M\times N}$ maps it to
\begin{equation}
  \mathbf{y}=\Phi\mathbf{z},
\end{equation}
and the sketched dictionary is $A=\Phi D$. The coefficient tensor
$X\in\R^{F\times|\mathcal{H}|\times 2}$ stores, for each candidate base frequency, all harmonic and quadrature coefficients. The recovery objective is
\begin{equation}
  \min_X
  \frac{1}{2}\|\mathbf{y}-\Phi D\,\operatorname{vec}(X)\|_2^2
  +\lambda\sum_{j=1}^{F}\|X_j\|_2.
  \label{eq:group_objective}
\end{equation}
Equation~\ref{eq:group_objective} connects the previous module to the solver: polynomial projection constructs $\mathbf{z}$, the dictionary $D$ encodes Eq.~\ref{eq:prop_signal}, and the group penalty asks the model to explain a window with only a few propeller candidates.

\subsubsection{Unfolded LISTA layers.}
Each layer performs a residual correction followed by harmonic-group soft thresholding:
\begin{equation}
  X^{k+1}=
  \eta_{\theta_k}\!\left(
  X^k + \alpha_k A^\top(\mathbf{y}-A\,\operatorname{vec}(X^k))
  \right).
  \label{eq:lista}
\end{equation}
The step size $\alpha_k$ and thresholds $\theta_k$ are learned through softplus-parameterized variables. By default, $\Phi$ is fixed and $D$ is learnable. This choice keeps the measurement path deterministic while allowing the harmonic atoms to adapt to real event-camera responses.

\subsubsection{Group shrinkage and spectrum.}
The shrinkage operator is applied to an entire frequency group:
\begin{equation}
  \eta_{\theta}(X_j) =
  \max\left(1-\frac{\theta_j}{\|X_j\|_2+\epsilon},0\right)X_j.
\end{equation}
Group-wise shrinkage keeps the learned layers tied to the harmonic recovery objective. The network learns a few steps to recover a compact harmonic explanation rather than arbitrary spatial templates.
After $K$ layers, \chg\ forms a group spectrum
\begin{equation}
  r_j=\|X^K_j\|_2,\quad j=1,\ldots,F.
\end{equation}

\begin{figure*}[t]
    \centering
    \begin{subfigure}[t]{0.27\textwidth}
        \centering
        \includegraphics[height=3.5cm]{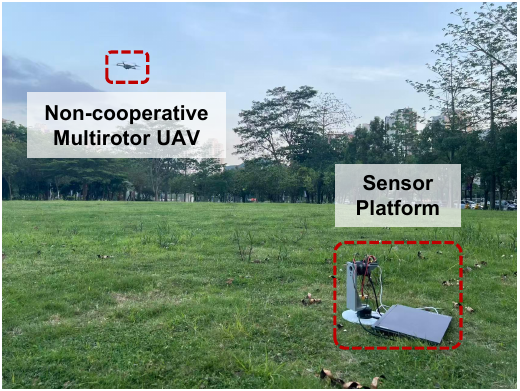}
        \caption{Real-time experimental scenario}
        \label{fig:real-time}
    \end{subfigure}
    \begin{subfigure}[t]{0.27\textwidth}
        \centering
        \includegraphics[height=3.5cm]{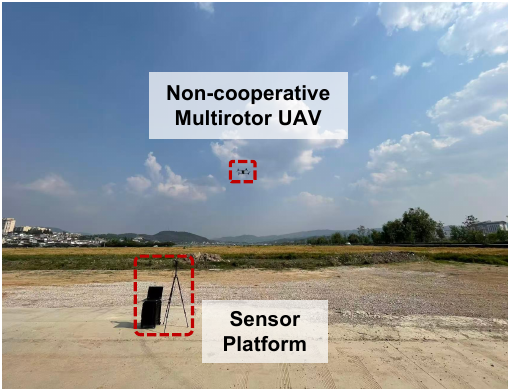}
        \caption{Long-range experimental scenario}
        \label{fig:long-range}
    \end{subfigure}
    \begin{subfigure}[t]{0.17\textwidth}
        \centering
        \includegraphics[height=3.5cm]{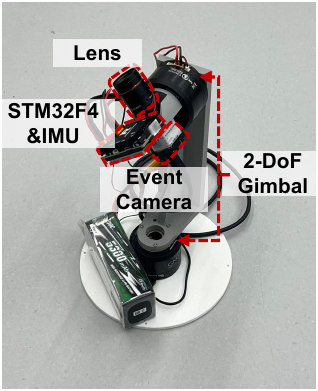}
        \caption{Sensor Platform}
        \label{fig:sensor}
    \end{subfigure}
    \begin{subfigure}[t]{0.27\textwidth}
        \centering
        \includegraphics[height=3.5cm]{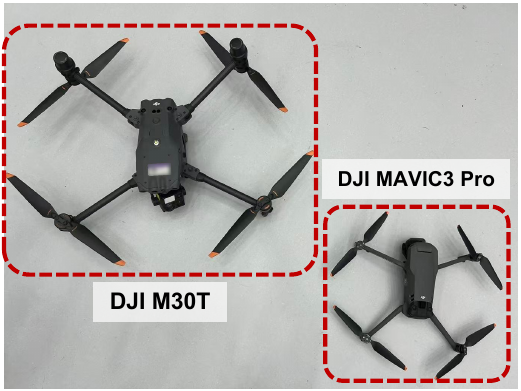}
        \caption{UAV}
        \label{fig:UAV}
    \end{subfigure}
    \vspace{0.2cm}
    \caption{Experimental setup of \sysname. The panels show the outdoor collection scene, sensing device, and target UAVs.}
    \Description{A compact four-panel experimental setup figure. Panel (a) shows the real-time experimental scenario. Panel (b) shows the long-range experimental scenario. Panel (c) shows the sensor platform. Panel (d) shows the UAV target.}
    \label{fig:deployment_scene}
\end{figure*}

\subsection{Local Peak Scoring and Candidate Evidence}
\label{sec:score_output}

The group spectrum recovered in Section~\ref{sec:lista} still has to be interpreted before it becomes candidate evidence. Its absolute scale changes with distance, sky region, scan speed, and event density, so a raw maximum of $r_j$ is not directly comparable across candidates or windows. 
\chg\ scores each peak relative to its local spectral floor and reports a candidate-level frequency-confidence pair. 
This keeps harmonic evidence available after spatial localization without using an early frequency gate to discard low-SNR targets.

\subsubsection{Local peak scoring.}
\label{sec:local_score}
A raw spectrum peak is brittle because local spectral floor can rise or fall with background activity and event density. \chg\ uses a local spectral-ratio score.

For each frequency bin $j$, the implementation excludes a small guard region around $j$ and averages neighboring reference bins:
\begin{equation}
  n_j =
  \frac{1}{|\mathcal{T}_j|}
  \sum_{\ell\in\mathcal{T}_j} r_\ell.
\end{equation}
It then reports the local ratio
\begin{equation}
  c_j = \frac{r_j}{n_j+\epsilon}.
\end{equation}
The verifier reports $\hat{j}=\argmax_j c_j,\hat{f}=f_{\hat{j}},\hat{c}=c_{\hat{j}}$.

\subsubsection{Peak frequency and confidence evidence.}
\sage\ provides the candidate support $\Omega_m$ before \chg. Given this support, \chg\ returns a peak score $\hat{c}_m$ and peak frequency $\hat{f}_m$. These values are kept as candidate-level harmonic evidence: they can support a candidate with stable harmonic structure, flag weak labels as ambiguous, or provide frequency metadata for later system modules. Proposal-time frequency gating is evaluated separately from the default pipeline so that candidate ranking and harmonic evidence recovery remain auditable.

\section{Implementation}
\label{sec:implementation}

\subsection{Algorithm Implementation}
\label{sec:algorithm_impl}

\textbf{Weak frequency labels.}
Long-range UAV recordings rarely contain synchronized motor telemetry. We therefore derive weak frequency labels from bbox annotations to indicate whether a candidate window contains the propeller-periodic cue used by \chg. These labels supervise training and support error audits; they are not online detector inputs and do not change the forward pass in Section~\ref{sec:chg_lista}.

\textbf{Label construction.}
For each annotated window, the labeler evaluates the target-region events against the local background. It accepts a label only when the frequency estimate has harmonic or base-frequency support, is separable from nearby background events, and remains stable across short sub-windows. Rejected windows keep a compact reason, such as weak evidence, harmonic ambiguity, background domination, or unstable temporal support.

\textbf{Training objective.}
The training path maps each synthetic or accepted weak label to the nearest candidate frequency bin and trains over the local spectral-ratio vector. The loss also preserves sketch-domain data consistency and group sparsity:
\begin{equation}
  \mathcal{L}
  =
  \mathcal{L}_{\mathrm{freq}}
  +\lambda_d\|\mathbf{y}-A\,\operatorname{vec}(X^K)\|_2^2
  +\lambda_s\frac{1}{F}\sum_j r_j.
\end{equation}
This keeps learning tied to a reconstructive harmonic explanation rather than only to a class label.

\textbf{Candidate-level harmonic evidence.}
At inference time, \sage\ provides candidate support $\Omega_m$, and \chg\ applies the recovery and local scoring path defined in Section~\ref{sec:score_output}. For each candidate, the verifier returns the peak frequency $\hat{f}_m$ and local spectral-ratio confidence $\hat{c}_m$. We treat this pair as candidate-level harmonic evidence: it indicates whether the localized support contains a compact propeller-frequency explanation. In the default inference path, this evidence is produced after candidate proposal and does not participate in bbox generation. Proposal-time frequency gating is evaluated only as a separate ablation.

\subsection{Platform Implementation}
\label{sec:platform_impl}

\textbf{Hardware platform.}
Figure~\ref{fig:deployment_scene} summarizes the real-world experimental setup, including two field scenes, the sensing device, and the UAV target. The real-time scene shows the outdoor prototype setting used to evaluate online sensing feasibility, while the long-range scene shows the far-distance collection setting used for detection and robustness experiments. The device view shows the event-camera gimbal platform, which consists of a Prophesee EVK4 HD event camera with an IMX636ES sensor, a long-focal-length lens, a RoboMaster Development Board Type C with an integrated BMI088 IMU, and a 2-Degree of Freedom (DoF) gimbal with yaw and pitch axes. The UAV view shows the target UAVs used in our experiments: the DJI M30T and DJI MAVIC 3 Pro Cine.

\section{Evaluation}
\label{sec:evaluation}

We evaluate \sysname\ in five parts. Section~\ref{sec:eval_methodology} defines the data, annotations, baselines, and metrics; Section~\ref{sec:eval_overall} reports long-range detection performance; Section~\ref{sec:eval_robustness} tests robustness across deployment conditions; Section~\ref{sec:eval_ablation} isolates the proposal-support and frequency-scoring choices; and Section~\ref{sec:eval_overhead} reports system overhead and real-time feasibility.

\begin{table*}[!t]
  \centering
  \caption{Range-balanced static-window detection comparison}
  \label{tab:eval_overall_detection}
  {\normalsize
  \renewcommand{\arraystretch}{1.15}
  \setlength{\tabcolsep}{1.3pt}
\begin{tabularx}{\textwidth}{@{}>{\raggedright\arraybackslash}p{0.125\textwidth}>{\raggedright\arraybackslash}p{0.145\textwidth}|*{6}{>{\centering\arraybackslash}X}|>{\centering\arraybackslash}p{0.060\textwidth}>{\centering\arraybackslash}p{0.090\textwidth}>{\centering\arraybackslash}p{0.060\textwidth}@{}}
\toprule
\rowcolor{black!10}\multicolumn{2}{c|}{\textbf{Method}} & \multicolumn{6}{c|}{\textbf{(a) Detection quality}} & \multicolumn{3}{c}{\textbf{(b) Error trade-off}} \\
\cmidrule(lr){1-2}\cmidrule(lr){3-8}\cmidrule(lr){9-11}
\rowcolor{black!4}\textbf{Method} & \textbf{Cue} & \textbf{P$_{.3}\uparrow$} & \textbf{R$_{.3}\uparrow$} & \textbf{F1$_{.3}\uparrow$} & \textbf{mAP$_{.3}\uparrow$} & \textbf{R$_{.5}\uparrow$} & \textbf{CHR$\uparrow$} & \textbf{FN$_{.3}\downarrow$} & \textbf{FP/win$_{.3}\downarrow$} & \textbf{CErr$\downarrow$} \\
\midrule
\textbf{EventRadar} & Harmonic & \textbf{0.911} & \textbf{0.991} & \textbf{0.949} & \textbf{0.990} & \textbf{0.876} & \textbf{1.000} & \textbf{0.009} & \textbf{0.098} & 2.24 \\
EV-SpSegNet & Point cloud & 0.814 & 0.813 & 0.814 & 0.778 & 0.702 & 0.991 & 0.187 & 0.182 & \textbf{2.22} \\
EvDetMAV & Propeller str. & 0.263 & 0.249 & 0.252 & 0.210 & 0.173 & 0.360 & 0.751 & 0.111 & 2.82 \\
EVPropNet & Sim. propeller & 0.110 & 0.169 & 0.130 & 0.038 & 0.062 & 0.311 & 0.831 & 1.751 & 3.32 \\
effqpdet & Density burst & 0.000 & 0.000 & 0.000 & 0.000 & 0.000 & 0.000 & -- & -- & -- \\
\bottomrule
\end{tabularx}

  }
\end{table*}

\begin{figure*}[!t]
  \centering
  \begin{minipage}[t]{0.242\textwidth}
    \centering
    \includegraphics[width=\linewidth]{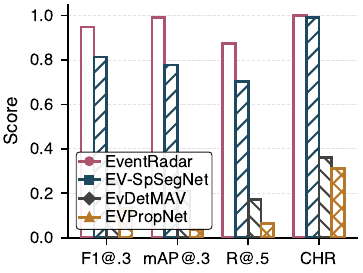}
    {\small (a) Balanced scores.\par}
  \end{minipage}\hfill
  \begin{minipage}[t]{0.242\textwidth}
    \centering
    \includegraphics[width=\linewidth]{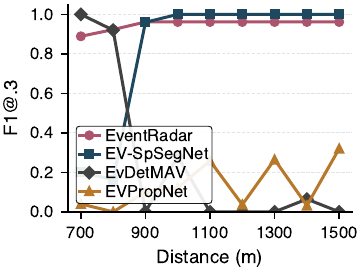}
    {\small (b) F1 vs. range.\par}
  \end{minipage}\hfill
  \begin{minipage}[t]{0.242\textwidth}
    \centering
    \includegraphics[width=\linewidth]{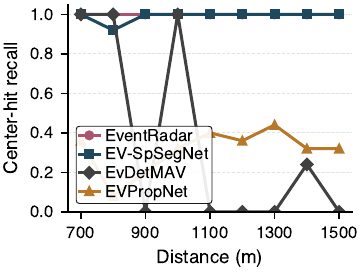}
    {\small (c) Center-hit recall.\par}
  \end{minipage}\hfill
  \begin{minipage}[t]{0.242\textwidth}
    \centering
    \includegraphics[width=\linewidth]{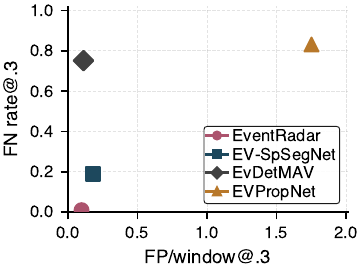}
    {\small (d) Error trade-off.\par}
  \end{minipage}
  \caption{Static-window detection trends}
  \Description{A four-panel quantitative figure comparing EventRadar, EV-SpSegNet, EVPropNet, and EvDetMAV on static long-range event-camera windows. EventRadar has the strongest balanced scores, keeps high F1 and center-hit recall across distance, and occupies the lowest missed-target and false-alarm trade-off region among non-degenerate event-camera detectors.}
  \label{fig:eval_overall_baseline_summary}
\end{figure*}

\subsection{Experimental Methodology}
\label{sec:eval_methodology}

\textbf{Field studies.}
We implement \sysname\ as a real-time long-range UAV sensing prototype and evaluate it in outdoor field studies. The field corpus contains more than 116\,GB of raw event recordings collected with our active observation platform, covering UAV observations from 700\,m to 1.5\,km across range, flight state, viewing angle, lighting, target speed, UAV model, and distractor conditions. For repeatable comparison, all methods are evaluated on recorded event streams replayed through the same sensing pipeline.

\textbf{Ground-truth annotations and protocols.}
For quantitative scoring, we build over 13k time-aligned bbox annotations on the field recordings. These annotations are used only after inference to compute detection and robustness metrics. We use stratified replay protocols rather than random frame sampling: the main detection benchmark is range-balanced, the robustness studies are condition-stratified factor audits, and the ablations are reported as distance-macro summaries. Frequency-related labels and diagnostics follow the weak-supervision procedure described in Section~\ref{sec:algorithm_impl}.

\textbf{Comparative methods.}
We compare \sysname\ with related event-camera baselines that cover three cue families. \\
\textit{(i) Event point-cloud methods:} EV-SpSegNet~\cite{ev_uav} treats the event stream as a sparse spatiotemporal point cloud and segments tiny UAV targets from point-cloud continuity. \\
\textit{(ii) Propeller spatial-structure methods:} EvDetMAV~\cite{evdetmav} detects rotary-wing MAVs from structured propeller-event patterns, while EVPropNet~\cite{evpropnet} trains on simulated propeller events and recognizes propeller morphology from event-derived spatial representations. \\
\textit{(iii) Event-density burst methods:} effqpdet~\cite{quad_prop_attr} detects rotating-propeller activity from local event-density bursts in high-resolution event streams. \\

\textbf{Metrics and evaluation scope.}
Detection quality is reported with precision, recall, F1, and mAP at IoU thresholds 0.3 and 0.5. Because protected-airspace response cueing is sensitive to missed targets and extra candidate boxes, we also report center-hit recall (CHR), false-negative rate (FN), false positives per target-present window (FP/win), image-plane center error (CErr), and normalized center error (NCE). CHR measures whether a predicted bbox center falls inside the target bbox, FN$_{.3}$ is $1-\mathrm{R}_{.3}$, FP/win counts IoU-unmatched predicted boxes normalized by the number of annotated replay windows, CErr is mean center error in pixels, and NCE divides the top-1 center error by the target-bbox diagonal.

\subsection{Overall Performance}
\label{sec:eval_overall}

\subsubsection{UAV Detection}
\label{sec:eval_detection_success}

Table~\ref{tab:eval_overall_detection} shows that \sysname\ delivers the strongest long-range event-camera bbox detection. On the target-present static-window benchmark, \sysname\ reaches 0.990 mAP$_{.3}$ and 0.949 F1$_{.3}$, while the strongest baseline, EV-SpSegNet, reaches 0.778 and 0.814, respectively. The larger difference appears in missed-target control: \sysname\ reduces FN$_{.3}$ to 0.009, over one order of magnitude lower than EV-SpSegNet.

This pattern is consistent with the cue used by each baseline. EV-SpSegNet exploits spatiotemporal continuity in sparse event point clouds, which is effective when the target trace is clear but becomes sensitive to fragmented long-range event support and background event structures. EvDetMAV and EVPropNet focus on propeller-local evidence, through event clustering or morphology learned from simulated events, so their outputs often remain partial propeller regions under the full-body bbox metric. effqpdet relies on local event-density bursts and produces no valid positive bbox in this long-range protocol. 

\subsubsection{Range-wise Stability}
\label{sec:eval_range_stability}

Figure~\ref{fig:eval_overall_baseline_summary} explains where the overall gain comes from. Under the static-window protocol, \sysname\ saturates CHR in every distance bin, so the detector keeps a usable target cue even when IoU-based box tightness varies. Its F1$_{.3}$ is 0.889 at 700\,m, 0.923 at 800\,m, and 0.962 from 900\,m to 1500\,m. EV-SpSegNet becomes highly accurate on the cleaner 900--1500\,m static bins, but its 700--800\,m recall and extra-box behavior reduce the macro score. EvDetMAV is strong on the nearest static bins but is not stable at longer range, and EVPropNet remains dominated by unmatched boxes. The stable center-hit and F1 trends indicate that \sysname\ can maintain reliable UAV cues at long range.

\subsubsection{Missed-target and False-alarm Trade-off}
\label{sec:eval_missed_false_alarm_tradeoff}

\begin{table*}[!t]
  \centering
  \caption{Flight-state robustness under range-stratified replay}
  \label{tab:eval_flight_state_robustness}
  {\normalsize
  \renewcommand{\arraystretch}{1.17}
  \setlength{\tabcolsep}{2.1pt}
  \setlength{\tabcolsep}{2.2pt}
\begin{tabularx}{\textwidth}{@{}l|*{4}{>{\centering\arraybackslash}X}|*{4}{>{\centering\arraybackslash}X}|*{4}{>{\centering\arraybackslash}X}@{}}
\toprule
\rowcolor{black!10}\multicolumn{1}{c|}{Method} & \multicolumn{4}{c|}{Static} & \multicolumn{4}{c|}{Translation} & \multicolumn{4}{c}{Spin} \\
\cmidrule(lr){1-1}\cmidrule(lr){2-5}\cmidrule(lr){6-9}\cmidrule(lr){10-13}
\rowcolor{black!4}Method & F1$_{.3}\uparrow$ & CHR$\uparrow$ & FN$_{.3}\downarrow$ & CErr$\downarrow$ & F1$_{.3}\uparrow$ & CHR$\uparrow$ & FN$_{.3}\downarrow$ & CErr$\downarrow$ & F1$_{.3}\uparrow$ & CHR$\uparrow$ & FN$_{.3}\downarrow$ & CErr$\downarrow$ \\
\midrule
\textbf{EventRadar} & \textbf{0.949} & \textbf{1.000} & \textbf{0.009} & 2.24 & \textbf{0.566} & \textbf{0.760} & \textbf{0.396} & \textbf{3.90} & \textbf{0.264} & 0.329 & \textbf{0.720} & 3.67 \\
EV-SpSegNet & 0.814 & 0.991 & 0.187 & \textbf{2.22} & 0.158 & 0.209 & 0.893 & 4.34 & 0.082 & 0.084 & 0.947 & 3.88 \\
EvDetMAV & 0.252 & 0.360 & 0.751 & 2.82 & 0.176 & 0.213 & 0.822 & 6.15 & 0.222 & 0.338 & 0.778 & \textbf{3.39} \\
EVPropNet & 0.130 & 0.311 & 0.831 & 3.32 & 0.075 & 0.422 & 0.898 & 10.27 & 0.090 & \textbf{0.364} & 0.893 & 6.41 \\
\bottomrule
\end{tabularx}

  }
\end{table*}

\begin{figure*}[!t]
  \centering
  \begin{minipage}[t]{0.242\textwidth}
    \centering
    \includegraphics[width=\linewidth]{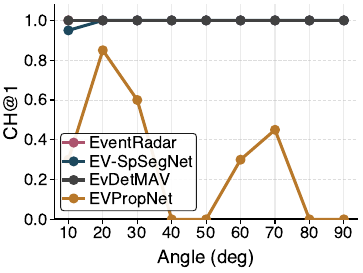}
    {\small (a) Angle: CH@1\par}
  \end{minipage}\hfill
  \begin{minipage}[t]{0.242\textwidth}
    \centering
    \includegraphics[width=\linewidth]{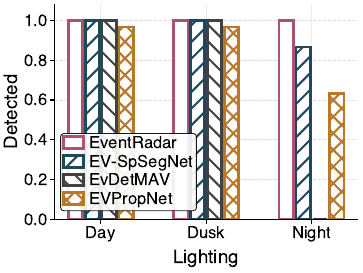}
    {\small (b) Lighting: detected\par}
  \end{minipage}\hfill
  \begin{minipage}[t]{0.242\textwidth}
    \centering
    \includegraphics[width=\linewidth]{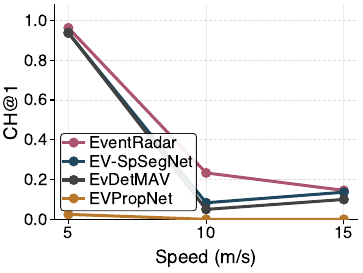}
    {\small (c) Speed: CH@1\par}
  \end{minipage}\hfill
  \begin{minipage}[t]{0.242\textwidth}
    \centering
    \includegraphics[width=\linewidth]{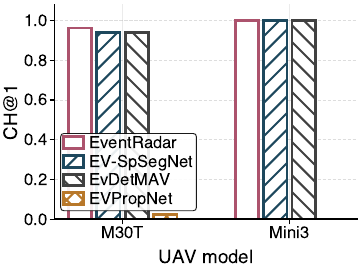}
    {\small (d) Model: CH@1\par}
  \end{minipage}

  \vspace{0.5ex}
  \begin{minipage}[t]{0.242\textwidth}
    \centering
    \includegraphics[width=\linewidth]{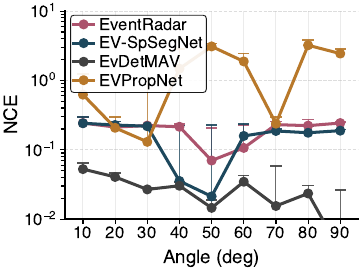}
    {\small (e) Angle: NCE\par}
  \end{minipage}\hfill
  \begin{minipage}[t]{0.242\textwidth}
    \centering
    \includegraphics[width=\linewidth]{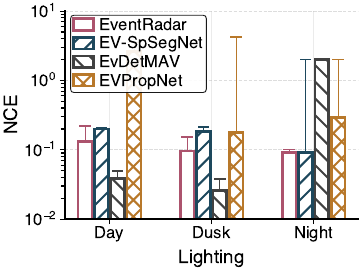}
    {\small (f) Lighting: NCE\par}
  \end{minipage}\hfill
  \begin{minipage}[t]{0.242\textwidth}
    \centering
    \includegraphics[width=\linewidth]{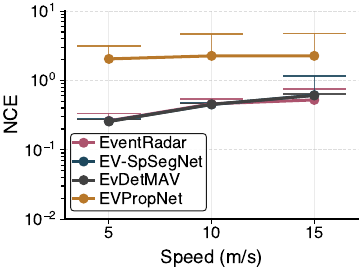}
    {\small (g) Speed: NCE\par}
  \end{minipage}\hfill
  \begin{minipage}[t]{0.242\textwidth}
    \centering
    \includegraphics[width=\linewidth]{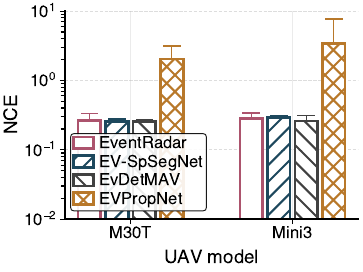}
    {\small (h) Model: NCE\par}
  \end{minipage}
  \caption{Condition-stratified robustness summary.}
  \Description{A two-row, four-column figure assembled from eight separate panel graphics comparing EventRadar, EV-SpSegNet, EvDetMAV, and EVPropNet under viewing angle, lighting, target speed, and UAV model changes. The top row reports center-hit or detection availability, and the bottom row reports normalized center error on logarithmic y-axes.}
  \label{fig:eval_factor_robustness}
\end{figure*}

Table~\ref{tab:eval_overall_detection}(b) and Figure~\ref{fig:eval_overall_baseline_summary}(d) evaluate the error trade-off most relevant to protected-airspace cueing: preserving target availability while limiting extra unmatched boxes in the annotated replay. \sysname\ gives the strongest balance, reducing FN$_{.3}$ to 0.009 while keeping FP/win$_{.3}$ at 0.098 under the same replay normalization. EV-SpSegNet localizes detected targets with a center error close to \sysname, but its missed-target rate and unmatched-box load are both higher. The propeller- and density-based baselines fail on one side of the trade-off: EvDetMAV misses many targets, EVPropNet produces a large unmatched-box load, and effqpdet produces no valid positive boxes. These patterns show that \sysname\ provides the most reliable long-range UAV cue among the evaluated event-camera baselines.

\subsection{Robustness Evaluation}
\label{sec:eval_robustness}

The robustness experiments vary one deployment factor at a time while keeping the detector, thresholds, and scan policy fixed. These condition-stratified audits report both detection quality and stability, because a robust anti-UAV system must avoid both missed detections and unstable boxes.

\subsubsection{Impact of UAV Flight State}
\label{sec:eval_flight_state}

We evaluate flight-state robustness on the 700--1500\,m long-range batch. Table~\ref{tab:eval_flight_state_robustness} reports the state-conditioned metrics. The audit groups annotated replay segments into three motion states: static or hovering, translational motion (3--5\,m/s UAV speed), and spin-candidate motion. All methods use the same label-centered matching protocol as the overall long-range detection benchmark.

Table~\ref{tab:eval_flight_state_robustness} shows that motion state is a much stronger stressor than range within the normal long-range batch. \sysname\ has the highest F1$_{.3}$ and the lowest FN$_{.3}$ in all three states. In the static subset, it reaches 0.949 F1$_{.3}$, saturates CHR, and keeps FN$_{.3}$ at 0.009, matching the controlled overall benchmark. Under translation, the score drops to 0.566 F1$_{.3}$ and 0.760 CHR, but it remains well above EV-SpSegNet, EvDetMAV, and EVPropNet in balanced detection quality. The cue families explain the main failure modes. EV-SpSegNet depends on point-cloud continuity in the event stream, so translation and spin fragment the target trace and sharply reduce recall. EvDetMAV and EVPropNet rely on propeller spatial structure, either through structured propeller-event patterns or morphology learned from simulated propeller events; these spatial cues become less stable when the target state changes from static support to translational or spin-candidate motion.

\subsubsection{Impact of Viewing Angle}
\label{sec:eval_viewing_angle}

We evaluate labeled viewing-angle sequences from 10$^\circ$ to 90$^\circ$ while keeping the detector and window length fixed. This factor study is designed to test whether the target remains findable under changing aspect.

Figure~\ref{fig:eval_factor_robustness}(a,e) summarizes the valid angle rows with CH@1 on the top row and NCE medians with P90 ticks on the bottom row. \sysname\ keeps full center-hit availability across all valid viewing angles. Its NCE median stays between 0.070 and 0.243, with the lowest error near the mid-angle 50$^\circ$ view. EV-SpSegNet and EvDetMAV also localize the center consistently in this controlled angle subset, while EVPropNet is unstable across aspect changes and fails at several high-angle views. These results show that \sysname\ maintains aspect-stable target availability and center localization for event-based target discovery.

\subsubsection{Impact of Lighting}
\label{sec:eval_lighting}

We audit lighting robustness on manually labeled event sequences under daytime, dusk, and night conditions. This subset is designed to test whether the target remains findable under changing event contrast, so Figure~\ref{fig:eval_factor_robustness}(b,f) reports detection availability together with top-1 normalized center error (NCE). For missed detections, the NCE value is penalized so that failures remain visible in the aggregate figure.

EventRadar keeps full detection availability across all three lighting conditions, with NCE medians of 0.131, 0.096, and 0.092 for daytime, dusk, and night, respectively. EvDetMAV is very accurate in daytime and dusk but produces no valid detection in the night subset, while EV-SpSegNet and EVPropNet show either missed detections or large error tails under low light. These results show that \sysname\ maintains lighting-stable target availability and center localization under changing event contrast.

\subsubsection{Impact of Target Speed}
\label{sec:eval_target_speed}

We evaluate speed-controlled DJI M30T sequences at 90$^\circ$, 100\,m, with target speeds of 5\,m/s, 10\,m/s, and 15\,m/s. All methods use the same center-localization protocol as the angle and lighting studies.

Figure~\ref{fig:eval_factor_robustness}(c,g) shows that higher target speed substantially reduces detection stability beyond the 5\,m/s matched setting. At 5\,m/s, \sysname\ reaches 0.963 CH@1 with an NCE median of 0.262. At 10 and 15\,m/s, CH@1 drops to 0.233 and 0.145. EV-SpSegNet and EvDetMAV remain competitive in NCE at 10\,m/s but collapse in top-1 center-hit rate, while EVPropNet remains dominated by false or off-target localizations. These results define a clear speed-related robustness boundary for the current prototype; under the 15\,m/s condition, \sysname\ still preserves the strongest loose-center success among the compared event-camera methods.

\subsubsection{Impact of UAV Size}
\label{sec:eval_uav_model}

We study whether detection quality changes with UAV size and apparent scale using matched low-speed rows: 5\,m/s, 90$^\circ$ view, and 100\,m range. This comparison includes DJI M30T and DJI MAVIC3 Pro windows under the same detector, scan policy, and center-localization protocol.

Figure~\ref{fig:eval_factor_robustness}(d,h) shows that \sysname\ remains stable across the two evaluated UAV sizes under matched slow motion. For \sysname, CH@1 is 0.963 on M30T and reaches full center-hit availability on MAVIC3 Pro, while the NCE medians stay in the 0.26--0.29 range. EV-SpSegNet and EvDetMAV are also strong in this low-speed setting, whereas EVPropNet is dominated by missed or false localizations. These results indicate that UAV size has limited impact on target availability in this matched setting.

\subsubsection{Kite Distractor Case Study}
\label{sec:eval_kite_case_study}

\begin{figure}[!t]
  \centering
  \begin{minipage}[t]{\columnwidth}
    \centering
    \includegraphics[width=\linewidth]{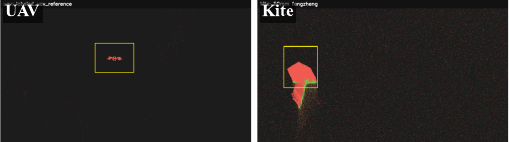}
    {\small (a) Event accumulation with ROI.\par}
  \end{minipage}

  \vspace{0.35ex}
  \begin{minipage}[t]{\columnwidth}
    \centering
    \includegraphics[width=\linewidth]{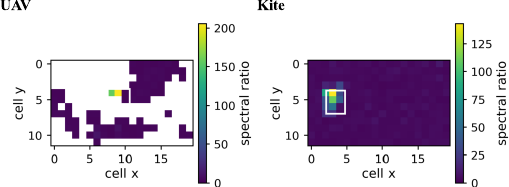}
    {\small (b) Local frequency-evidence map.\par}
  \end{minipage}

  \vspace{0.35ex}
  \begin{minipage}[t]{\columnwidth}
    \centering
    \includegraphics[width=\linewidth]{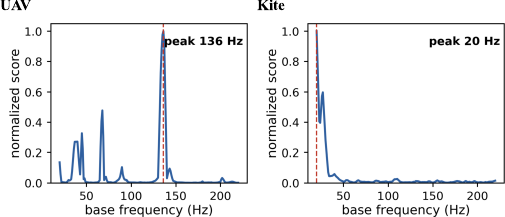}
    {\small (c) ROI harmonic spectrum.\par}
  \end{minipage}
  \caption{Qualitative UAV--kite frequency evidence.}
  \Description{A compact single-column figure assembled from three independent panel graphics. Each panel compares a labeled UAV reference with a density-screened kite window. The panels show short-window event accumulation with ROI, local frequency-evidence map, and ROI harmonic spectrum.}
  \label{fig:eval_kite_case_study}
\end{figure}

\begin{figure*}[!t]
  \centering
  \begin{minipage}[t]{0.30\linewidth}
    \centering
    \includegraphics[width=\linewidth]{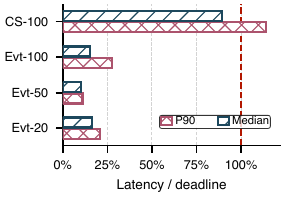}
    {\small (a) Online trace.\par}
  \end{minipage}\hfill
  \begin{minipage}[t]{0.30\linewidth}
    \centering
    \includegraphics[width=\linewidth]{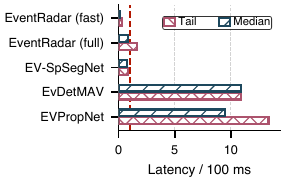}
    {\small (b) Runtime audit.\par}
  \end{minipage}\hfill
  \begin{minipage}[t]{0.30\linewidth}
    \centering
    \includegraphics[width=\linewidth]{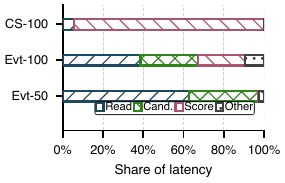}
    {\small (c) Breakdown.\par}
  \end{minipage}
  \caption{System overhead and real-time feasibility of the EventRadar prototype.}
  \Description{Three panels summarize online path slack, detector-runtime audit, and online latency breakdown.}
  \label{fig:eval_overhead}
\end{figure*}

We use the kite data only as a qualitative distractor case study. Candidate kite windows are screened from the pure-kite recording.  

Figure~\ref{fig:eval_kite_case_study} illustrates the intended claim boundary. The UAV reference has a compact target ROI and a strong, narrow rotor-band harmonic response, with a 136\,Hz peak and a local spectral ratio of 255.6. The kite window is visually obvious and produces strong event support from the kite body, but its ROI spectrum is concentrated at the low edge of the searched band, with a 20\,Hz peak and a lower local ratio of 130.2. We therefore use this example only to show that visually active non-UAV objects can generate event support whose frequency evidence differs qualitatively from the UAV reference.

\subsection{Controlled Module Ablation}
\label{sec:eval_ablation}

We use controlled module-replacement ablations for the two interfaces in \sysname. In each test, the surrounding protocol is fixed and only one module output is replaced: proposal-side local support before temporal readout, or candidate-level scoring on a fixed proposal pool.

\subsubsection{Proposal-Side Support}
\label{sec:eval_ablation_proposal_support}

The proposal-side ablation keeps the downstream evaluation protocol fixed and changes how local event support is formed before \chg. \textit{Fixed tile} scans uniform 32$\times$32 image tiles and scores them with a frame-style classifier. \textit{Frame adaptive} forms connected components on an accumulated event frame. \textit{Adaptive support} uses the local event-support proposal described in Section~\ref{sec:background_aware_candidate}, preserving compact event clusters as candidate regions for later temporal readout.

\subsubsection{Frequency Scoring}
\label{sec:eval_ablation_frequency_scoring}

The frequency-scoring ablation freezes the adaptive-support candidate pool and replaces only the scalar score assigned to each candidate. \textit{Event count} uses raw event density. \textit{FFT peak} uses the strongest spectral peak. \textit{CS score} uses the compact-sensing reconstruction score. \textit{\chg\ score} applies the harmonic-group scorer from Section~\ref{sec:chg_lista}.

\begin{table}[!t]
  \centering
  \normalsize
  \renewcommand{\arraystretch}{1.17}
  \setlength{\tabcolsep}{1.4pt}
  \caption{Controlled proposal-support and frequency-scoring ablations.}
  \label{tab:eval_ablation_controlled}
  \begin{tabularx}{\columnwidth}{@{}l|*{6}{>{\centering\arraybackslash}X}@{}}
    \toprule
    \rowcolor{black!10}Proposal & R$_{.3}\uparrow$ & AP$_{.3}\uparrow$ & R$_{.5}\uparrow$ & AP$_{.5}\uparrow$ & F1$_{.3}\uparrow$ & CErr$\downarrow$ \\
    \midrule
    Fixed tile & 0.116 & 0.020 & 0.017 & 0.002 & 0.105 & 9.02 \\
    Frame adaptive & 0.766 & 0.764 & 0.522 & 0.404 & \textbf{0.682} & \textbf{2.19} \\
    \textbf{Adaptive support} & \textbf{0.911} & \textbf{0.905} & \textbf{0.700} & \textbf{0.571} & 0.456 & 2.62 \\
    \bottomrule
  \end{tabularx}

  \vspace{0.9ex}
  \begin{tabularx}{\columnwidth}{@{}l|*{6}{>{\centering\arraybackslash}X}@{}}
    \toprule
    \rowcolor{black!10}Scorer & Top1$_{.3}\uparrow$ & AP$_{.3}\uparrow$ & mAP$_{.3}\uparrow$ & R$_{.3}\uparrow$ & F1$_{.3}\uparrow$ & mAP$_{.5}\uparrow$ \\
    \midrule
    Event count & 0.088 & 0.165 & 0.064 & 0.149 & 0.074 & 0.034 \\
    FFT peak & 0.630 & 0.663 & 0.629 & 0.663 & 0.331 & 0.418 \\
    CS score & 0.856 & 0.865 & 0.853 & 0.866 & 0.433 & 0.488 \\
    \textbf{\chg\ score} & \textbf{0.928} & \textbf{0.928} & \textbf{0.920} & \textbf{0.928} & \textbf{0.464} & \textbf{0.505} \\
    \bottomrule
  \end{tabularx}
\end{table}

In the compact headers, AP denotes mAP for the proposal-support ablation and candidate AP for the frequency-scoring ablation. Table~\ref{tab:eval_ablation_controlled} shows that fixed tiling is too coarse for kilometer-range UAVs, while adaptive support preserves the highest recall and AP at both IoU thresholds. With the same candidate pool fixed, \chg\ reaches 0.928 Top1$_{.3}$ and 0.920 mAP$_{.3}$, ahead of CS, FFT peak, and raw event count. These replacements support the deployed boundary: \sage\ preserves localized event candidates, and \chg\ supplies the candidate-level harmonic evidence used to select among them.

\subsection{System Overhead and Real-Time Feasibility}
\label{sec:eval_overhead}

Because \sysname\ targets a scanning prototype, we profile the Jetson Orin implementation on the same online path used by the prototype: events are replayed in timestamp order, windowed by the scan-loop stride, and converted to bbox predictions before the next window arrives. Figure~\ref{fig:eval_overhead} reports latency relative to the window deadline, an implementation breakdown, and a detector-runtime audit. In the figure labels, CS denotes the compact-sensing bbox path, Evt denotes the fast event-candidate path, and Cand. denotes candidate generation. The audit uses two input protocols: cached 100\,ms event windows aligned with Sections~\ref{sec:eval_overall}--\ref{sec:eval_robustness} for fair detector comparison, and a continuous timestamp-ordered replay trace for scan-loop feasibility.

Figure~\ref{fig:eval_overhead}(a) shows that the fast event-candidate path leaves a large scan-loop margin on Jetson Orin, while the compact-sensing bbox path is accurate but close to the 100\,ms deadline. Figure~\ref{fig:eval_overhead}(b) shows the system-level timing boundary: EventRadar's fast path stays comfortably real-time, the full path is near the deadline, and the pinned original EvDetMAV and EVPropNet rows are too slow under their recorded runtime protocols. Figure~\ref{fig:eval_overhead}(c) explains this gap: GPU Multiple Measurement Vector-Orthogonal Matching Pursuit (MMV-OMP) scoring dominates the compact-sensing path, whereas event-support ranking keeps the fast online candidate path lightweight.

\section{Conclusion}
\label{sec:conclusion}

\sysname\ reframes protected-airspace UAV detection as long-range visual UAV discovery under wide-area active sensing. The central idea is simple: propeller-induced temporal periodicity can remain detectable after spatial morphology becomes unreliable. The system therefore uses \sage\ to distinguish tiny moving targets from persistent backgrounds via spatiotemporal correlation, store candidate event clusters in a bearing‑indexed scene memory, and then apply \chg\ to recover temporal harmonic evidence from those clusters through a harmonic‑evidence module. In this way, UAV identification is achieved simultaneously over long‑range and wide‑area operations.

\begingroup
\setlength{\bibsep}{1pt plus 0.2pt}
\renewcommand{\bibfont}{\scriptsize}
\bibliographystyle{ACM-Reference-Format}
\bibliography{sample-base}
\endgroup

\end{document}